\documentclass[runningheads,a4paper]{llncs}

\usepackage{amssymb}
\usepackage{graphicx}

\usepackage{graphicx}
\usepackage{epstopdf}

\usepackage{booktabs}

\usepackage{verbatim}
\usepackage{algorithm}
\usepackage{algpseudocode}
\usepackage{xspace}
\usepackage{amsmath}
\usepackage{caption}
\usepackage{subcaption}

\usepackage{rotating}
\usepackage{verbatimbox}
\usepackage[table,xcdraw]{xcolor}
\usepackage{multirow}
\usepackage{amsmath}
\usepackage{graphicx}
\usepackage[colorinlistoftodos]{todonotes}
\usepackage[colorlinks=true, allcolors=blue]{hyperref}
\usepackage{cite}
\usepackage[overlay,absolute]{textpos}

\begin{document}

\begin{textblock*}{10in}(30mm, 15mm)
{\textbf{Ref:} \emph{International Conference on Artificial Neural Networks (ICANN)}, Springer LNCS,}
\end{textblock*}
\begin{textblock*}{10in}(30mm, 20mm)
{Vol.~11141, pp.~613--621, Rhodes, Greece, October 2018.}
\end{textblock*}

\mainmatter

\title{Handwriting-Based Gender Classification \\
Using End-to-End Deep Neural Networks}

\titlerunning{Handwriting-Based Gender Classification Using Deep Neural Networks}

\author{Evyatar Illouz\inst{1} \and Eli (Omid) David\inst{1} \and Nathan S. Netanyahu\inst{1,2} }

\authorrunning{E. Illouz, E.O. David, and N.S. Netanyahu}

\institute{
Department of Computer Science, Bar-Ilan University, Ramat-Gan 52900, Israel \\
\email{iluz101@gmail.com, mail@elidavid.com, nathan@cs.biu.ac.il}
\and
Center for Automation Research, University of Maryland, College Park, MD 20742\\
\email{nathan@cfar.umd.edu}
}

\maketitle

\begin{abstract}

Handwriting-based gender classification is a well-researched problem that has been approached mainly by traditional machine learning techniques. In this paper, we propose a novel deep learning-based approach for this task. Specifically, we present a \textit {convolutional neural network} (CNN), which performs automatic feature extraction from a given handwritten image, followed by classification of the writer’s gender. Also, we introduce a new dataset of labeled handwritten samples, in Hebrew and English, of 405 participants. Comparing the gender classification accuracy on this dataset against human examiners, our results show that the proposed deep learning-based approach is substantially more accurate than that of humans.

\end{abstract}

\section{Introduction}

Gender classification by handwriting is a well-studied problem, assuming that one's gender can be predicted based on their handwriting. Although there has been a considerable amount of research on this subject, it is still considered a challenging problem. In fact, neither computerized analysis nor humans, have achieved highly-accurate results for this task, as of yet.

The common assumption is that various demographic properties can be learned by studying the discriminative features of a person's handwriting, e.g., gender, handedness (i.e., whether the person is left-/right-handed), age bracket, ethnicity, etc. Indeed, human handwriting is used to examine and investigate human characteristics in a variety of applications, such as mail sorting~\cite{bouadjenek2014local}, bank check verification~\cite{bandi2005writer,bouadjenek2014local}, personality profiling~\cite{shackleton1994european,king2000illusory}, historical document analysis~\cite{ahmed2017improving}, and criminological/forensic investigations~\cite{bouadjenek2014local,bouadjenek2015age}.

Most of the recent approaches to gender classification by handwriting have evolved mainly around the same few datasets, i.e., the training and testing of these methods have been confined typically to a handful of datasets, such as the IAM on-line~\cite{IAMOnDB}, QUWI~\cite{al2012quwi}, KHATT~\cite{mahmoud2014khatt}, and MSHD~\cite{djeddi2014lamis} datasets. The motivation in this paper is mainly twofold: (1) Propose an improved gender classification method, and (2) augment the current pool of handwriting datasets in a significant manner. Specifically, we propose a new \textit{convolutional neural network} (CNN) variant for the gender classification task, which is relatively simple, efficient, and accurate. Also, we present a fairly large and diverse dataset, the \textit{Hebrew-English Bar-Ilan University} (HEBIU) offline handwriting dataset, which consists of 810 Hebrew and English handwriting samples, collected from a group of 405 participants. The newly-generated dataset would allow for extended research and comparative studies, regarding the classification of various attributes of interest.
Our results are comparable to those reported by previous methods, and they are substantially better than the accuracy rates obtained by human examiners on our HEBIU dataset.

\section{Related Work} \label{relatedWork}
Several machine learning techniques have been applied during the past two decades to the handwriting gender classification task. These approaches are based typically on feature extraction and training classifier; see Table~\ref{resultsComparison} below (extended from Gattal et al.~\cite{gattal2018gender}), for an overview.

% \usepackage{booktabs}
% \usepackage{graphicx}
\begin{table}[htbp]
\caption{Overview of handwriting gender classification techniques.}
\label{resultsComparison}
\begin{center}
\resizebox{\textwidth}{!}{%
\begin{tabular}{|c|c|c|c|c|}
\hline
\textbf{Research} & \textbf{Features} & \textbf{Classifier} & \textbf{Dataset} & \textbf{Accuracy}\\
\hline
\hline
Cha et al.~\cite{cha2001priori} (2001) & 
A set of macro and micro features & ANN & 
CEDAR~\cite{hull1994database} & 70.20\% \\ \hline

Liwicki et al.~\cite{liwicki2007automatic} (2011) & 
Combination of online \&  offline features  & GMM & 
IAM-OnDB~\cite{IAMOnDB}  & 65.57\% \\ \hline

Youssef et al.~\cite{youssef2013automated} (2013) & Gradient \& WD-LBP & SVM & QUWI~\cite{al2012quwi} & 74.30\% \\ \hline

Al-Maadeed et al.~\cite{al2014automatic} (2014) & Geometric & Random forests & QUWI~\cite{al2012quwi} & 73\% \\ \hline

Bouadjenek et al.~\cite{bouadjenek2014local} (2014) & HoG \& LBP & SVM & IAM-OnDB~\cite{IAMOnDB} & 74\% \\ \hline

Siddiqi et al.~\cite{siddiqi2015automatic} (2015) & Orientation curvature \& legibility & SVM & QUWI~\cite{al2012quwi} \& MSHD~\cite{djeddi2014lamis} & 68.75\%/73.02\% \\ \hline

Mirza et al.~\cite{mirza2016gender} (2016) & Gabor filters \& Fourier transform & ANN & QUWI~\cite{al2012quwi} & 70\% \\ \hline

Akbari et al.~\cite{akbari2017wavelet} (2017) & Wavelet sub-hands & SVM/ANN & QUWI~\cite{al2012quwi} \& MSHD~\cite{djeddi2014lamis} & 80\% \\ \hline

Ahmed et al.~\cite{ahmed2017improving} (2017) & Textural & Ensemble of classifiers & QUWI~\cite{al2012quwi} & 79\%--85\% \\ \hline

Gattal et al.~\cite{gattal2018gender} (2018) & Oriented basic image features & SVM & QUWI~\cite{al2012quwi} & 68\%--76\% \\ \hline

Morera et al.~\cite{morera2018gender} (2018) & Word seperation & CNN & IAM~\cite{IAMOnDB} \& KHATT~\cite{mahmoud2014khatt} & 80.72\%/68.9\% 
\\
\hline
\end{tabular}
}
\end{center}
\end{table}
%------------------------------

Cha et al.~\cite{cha2001priori} trained an \textit{artificial neural network} (ANN) in order to classify demographic sub-categories (such as gender, handedness, and age group) by using their own uppercase letter dataset. Later, they extended their work~\cite{bandi2005writer} to train a feed-forward neural network for feature extraction and classification, using enhancement techniques as bagging and boosting.
Their improved gender classifier achieved an accuracy rate of 77.5\% using 800 writing samples for training and 400 samples for testing.

Liwicki et al.~\cite{liwicki2007automatic} applied \textit{support vector machines} (SVM) and \textit{Gaussian mixture models} (GMM) to gender classification on the IAM-OnDB handwriting dataset. Their classifier achieved accuracy rates of 62\% and 67\%, respectively, using SVM and GMM.

Youssef et al.~\cite{youssef2013automated} proposed using \textit{wavelet domain local binary patterns} (WD-LBP) to train several SVM classifiers on both English and Arabic handwritings. Their classifier achieved an accuracy rate of 74.3\% on (a subset of) the QUWI dataset. 

Al-Maadeed et al.~\cite{al2014automatic} proposed using geometric features to classify age, gender, and nationality. Their proposed method applies \textit{random forests} and \textit{kernel discriminant analysis} for both \textit{text-dependent} and \textit{text-independent} classifications (i.e., same/different texts, respectively, of different writers are used for training and testing).
Their classifier achieved an overall accuracy of 73\% on the QUWI dataset.

Bouadjenek et al.~\cite{bouadjenek2014local} proposed extracting local descriptors, such as \textit{histogram of oriented gradients} (HoG), \textit{local binary patterns} (LBP), and grid features for offline handwriting, and then classifying them by SVM. Their method achieved an accuracy rate of 74\% on the IAM offline dataset. Likewise, Bouadjenek et al.~\cite{bouadjenek2015age} used local descriptors, such as \textit{gradient local binary patterns} (GLBP) and HoG to train an SVM classifier
to predict age, gender, and handedness. Their classifier achieved accuracy rates in the range of 69\%--74\% on the IAM-OnDB and KHATT datasets.

Similarly, Siddiqi et al.~\cite{siddiqi2015automatic} enhanced handwriting features by computing local and global features (e.g., inclination, texture, curvature, legibility, etc.), which are then used in ANN and SVM classifiers to distinguish between genders. Their classifier achieved accuracy rates of 68.75\% and 73.02\%, respectively, on the QUWI and MSHD datasets.

Mirza et al.~\cite{mirza2016gender} concentrated on the visual appearance of handwriting to investigate its effect on a writer's gender. They extract textural information by applying a bank of \textit{Gabor filters} to handwriting images from the QUWI dataset. They then use the mean and standard deviation of each handwriting plus its Fourier transform as input features for a feed-forward neural network. Their classifier achieved an accuracy rate of 70\% on the QUWI dataset.

Akbari et al.~\cite{akbari2017wavelet} extracted a feature vector based on a series of wavelet sub-bands quantized to produce a \textit{probabilistic finite state automaton}. This feature vector is then used to train ANN and SVM classifiers on the QUWI and MSHD datasets, and perform text-dependent and text-independent, as well as \textit{script-dependent} and \textit{script-independent} classifications (i.e., same/different languages, respectively, used for training and testing). They also introduced cross-database evaluations.

To enhance accuracy rates on the gender task,
Ahmed et al.~\cite{ahmed2017improving} used bagging, voting, and stacking of various classifiers based on some of the textural features mentioned earlier. They achieved accuracy rates in the range of 79\%--85\% on (a subset of) the QUWI dataset.

Gattal et al.~\cite{gattal2018gender} proposed using textural information from handwriting as the discriminative attribute between genders. They used image binarization and \textit{oriented basic image features}.
Their classifier achieved accuracy rates of 71\%, 76\%, and 68\% on the QUWI dataset, according to the protocols of  ICDAR 2013, ICDAR 2015, and ICFHR 2016, respectively.

Finally, Morera et al.~\cite{morera2018gender} were the first to apply a deep CNN for classifying a writer's demographics. They proposed the same architecture for both gender and handedness, 
as well as an architecture for the combined 4-class problem. Their gender classifier
achieved accuracy rates of 80.72\% and 68.9\%, respectively, on the IAM-OnDB and KHATT datasets.

To summarize, most of the surveyed methods exploit knowledge about the domain to extract certain features from the above datasets, and then train a machine learning module to classify these extracted features. In contrast, we present in this work a deep learning module, which performs essentially \textit{automated} feature extraction and classification, in a rather simple and efficient manner (requires no tedious preprocessing, and is far less complex than the system reported, e.g., by
Morera et al.~\cite{morera2018gender}).

\section{Proposed Method} \label{proposedMethod}
%note-note%
\subsection{The HEBIU Offline Handwriting Dataset}
Our newly generated dataset, the Hebrew-English Bar-Ilan University (HEBIU) offline handwriting dataset, contains 810 Hebrew and English handwriting samples of 405 participants from Israel. Each participant received a standard form, and was asked to write certain texts in Hebrew and English without any writing restrictions (e.g., pen type, pressure, etc.). In addition, each contributer was asked to provide personal data, such as gender, age, height, handedness, native language, country of birth, religion, education level, and profession.

Each such form was scanned by a 300dpi HP OfficeJet Pro 8710, in color
mode and JPEG format, at a high resolution of $2480 \times 3504$. 

The added value of our newly presented HEBIU dataset lies in the fact that it contains (also) hundreds of labeled writing samples in Hebrew, as well as diverse personal information per each participant. Thus, additional tasks, such as writer identification/verification and the classification of various demographic characteristics from handwriting samples, can be further pursued with such data.

\subsection{Handwriting Preprocessing}
As previously mentioned, our HEBIU dataset contains 810 Hebrew and English handwriting samples of 214 males and 191 females (i.e., of a total of 405 participants). Thus, to keep the data balanced, we excluded from the dataset, as part of preprocessing, 23 of the male forms.

\begin{figure}
\centering
         \begin{subfigure}[t]{0.4\textwidth}
                \centering
                \fbox{\includegraphics[height=1.8cm, width=1.8cm]{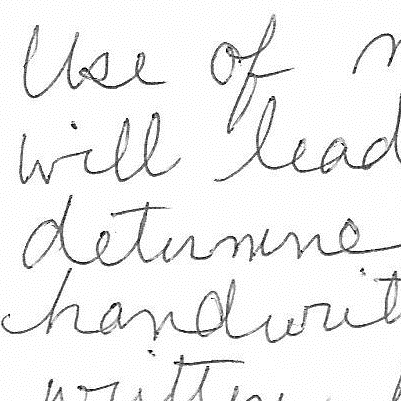}}
                \caption{}
                \label{fig:engSection}
        \end{subfigure}
        ~
		\vspace{0.2cm}
        \hspace{0.5cm}
        \begin{subfigure}[t]{0.4\textwidth}
                \centering
                \fbox{\includegraphics[height=1.8cm, width=1.8cm]{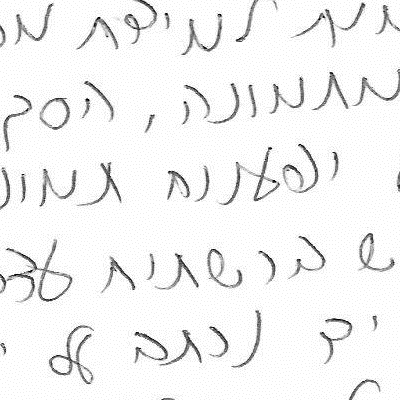}}
                \caption{}
                \label{fig:hebSection}
        \end{subfigure}

         \begin{subfigure}[t]{0.4\textwidth}
                \centering
                \fbox{\includegraphics[height=0.6cm, width=1.8cm]{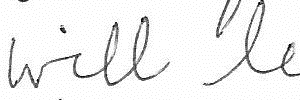}}
                \caption{}
                \label{fig:hebSection}
        \end{subfigure}
        ~
        \hspace{0.5cm}
        \begin{subfigure}[t]{0.4\textwidth}
                \centering
                \fbox{\includegraphics[height=0.6cm, width=1.8cm]{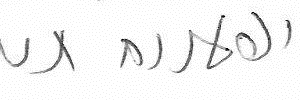}}
                \caption{}
                \label{fig:hebSection}
        \end{subfigure}
        
        \caption{Examples of resized text patches: (a)+(b) $100 \times 100$ English and Hebrew squared patches, and (c)+(d) $30 \times 100$ English and Hebrew rectangular patches.}
        \label{fig:patches_examples}
\end{figure}

In addition, the data should be normalized to be compatible with the network's architecture. Therefore, the first step was to extract a portion of the page which contains handwritten text, and convert it to a grayscale image. Afterwards, in order to enhance our data, we generated $N$ random patches for each form, of size $K \times M$, with (possible) overlaps between patches. A patch can be either a square or a rectangle. A square patch is meant to extract a whole subsection of words, while a rectangular patch is used to extract a line of text (or part of it), a single word, a writing sequence, etc. Both cases are illustrated in Fig.~\ref{fig:patches_examples}.

Having experimented extensively with the number of patches, as well as patch types and patch sizes, we converged eventually on $N = 200$ patches per handwritten sample and squared patches of size $400 \times 400$ pixels (i.e., $K = M = 400$). To keep the computational effort feasible, the patches were downscaled by 75\% to $100 \times 100$ pixels. (Similarly, the originally extracted rectangular patches of size $150 \times 500$ were downscaled to $30 \times 100$.)

Naturally, some of the generated patches were blank or contained small amounts of data. To overcome the selection of sparse text patches, we conducted a series of experiments to determine a threshold, based on a minimum ratio between black pixels and the total amount of pixels in a given patch. This was then used to select patches which contained a sufficient amount of data. 
Note that eventually we extracted 200 valid patches per each form.

\subsection{Network Architecture}

Our proposed network architecture is a CNN variant which inputs a grayscale, $100 \times 100$ patch and outputs the gender prediction. It is comprised of a total of four convolutional layers, followed by a single fully-connected layer and a softmax output layer, where all of the filters used are of size $3 \times 3$. More precisely, the first two layers consist of 64 and 128 filters, respectively, followed by a max pooling layer of $2 \times 2$ with a dropout of 0.4. The next two layers have the same structure, followed by a $2 \times 2$ max pooling layer with a dropout of 0.6.
Finally, a fully-connected layer with 128 neurons was added with a dropout of 0.5. The following network's hyper-parameters were picked: 20 epochs, a \textit{rectified linear unit} (ReLu) activation function~\cite{nair2010rectified}, an \textit{Adadelta} optimizer, 
and a binary cross entropy loss function.

\subsection{Accuracy Evaluation by Patch Aggregation}
We considered the following two classification measures, for a given handwriting sample:
\begin{enumerate}
\item \textit{Majority vote}: The gender class is determined based on the majority of classified patches, where the classification of each patch depends on whether the corresponding softmax value exceeds 0.5.
\item \textit{Average softmax}: The form is classified according to the average softmax value over the form's 200 patches.
\end{enumerate}

\section{Experimental Results} \label{results}
We divided the gender classification problem, in the context of this work, into three main types: (1) \textit{Intra-language} classification, where training and testing are conducted on the same language, (2) \textit{inter-language} classification, where training is conducted on one language and testing on the other, and (3) \textit{mixed language} classification, where both training and testing are conducted on both languages. For each type, we ran a 10-fold cross validation as follows. A fixed 20\% of the data (i.e., the same 76 forms) were set aside for testing, and 70\% (i.e., 268 forms) and 10\% (i.e., 38 forms) of the data, respectively, were allocated at random (from the remaining 80\%) for training and validation.

\subsection{Intra-Language Classification}
Regarding intra-language classification, we obtained average accuracy rates of 73.02\% and 75.26\%, respectively, in the case of Hebrew-Hebrew (i.e., training and testing performed on Hebrew texts) and English-English (i.e.,  both training and testing done on English texts).

\subsection{Inter-Language Classification}
For inter-language classification, we achieved accuracy rates of 75.65\% and 58.29\%, respectively, in the case of Hebrew-English classification (i.e., training on a Hebrew handwriting and testing on an English one) and  English-Hebrew classification (i.e., training on an English handwriting and testing on a Hebrew one). 

One attempt to explain this anomaly might be that since English is a second natural language in Israel (after Hebrew), the discriminative features between gender handwritings are less prominent (than in Hebrew), so generalizing becomes more challenging.

\subsection{Mixed Language Classification}
Enhancing our data by combining the texts of both languages yields an overall test accuracy of 77\% for both languages; in particular, 74.61\% and 79.34\% accuracy rates when tested on Hebrew and English texts, respectively.

\subsection{Summary of Results}
Table \ref{KCrossMinMaxResultsTB} summarizes the results, providing average accuracy rates and standard deviations for each method.
% \usepackage{booktabs}
% \usepackage{multirow}
% \usepackage{graphicx}

%------------------------------
\begin{table}[htbp]
\caption{Accuracy for gender classification types with 10-fold cross-validation (``HE'' stands for Hebrew, and ``EN'' stands for English).}
\label{KCrossMinMaxResultsTB}
\begin{center}
\resizebox{\textwidth}{!}{%
\begin{tabular}{|c|c|c|c|c|c|c|c|}
\hline
\multirow{2}{*}{\textbf{Experiment}} & \multirow{2}{*}{\textbf{Train}} & \multirow{2}{*}{\textbf{Test}} & \textbf{Accuracy} & \multirow{2}{*}{\textbf{Avg}} & \textbf{Std} & \textbf{Min} & \textbf{Max}\\
 &  &  & \textbf{Method} &  & \textbf{Dev} & \textbf{Accuracy} & \textbf{Accuracy}\\
\hline
\hline
\multirow{4}{*}{Intra-Language} & \multirow{2}{*}{HE} & \multirow{2}{*}{HE} & Majority vote 
& 73.02\% & 2.42 & 67.10\% & 75.00\% \\
& & & Avg. softmax 
& 72.89\% & 2.34 & 67.10\% & 75.00\% \\
\cline{2-8}
& \multirow{2}{*}{EN} & \multirow{2}{*}{EN} & Majority vote 
& 74.47\% & 2.65 & 69.74\% & 77.63\% \\
& & & Avg. softmax 
& 75.26\% & 2.47 & 71.05\% & 77.63\% \\ 
\hline

\multirow{4}{*}{Inter-Language} & \multirow{2}{*}{HE} & \multirow{2}{*}{EN} & Majority vote 
& 75.52\% & 6.86 & 60.52\% & 82.89\% \\
& & & Avg. softmax 
& 75.65\% & 7.40 & 57.89\% & 82.89\% \\
\cline{2-8}
& \multirow{2}{*}{EN} & \multirow{2}{*}{HE} & Majority vote 
& 58.29\% & 5.89 & 48.68\% & 65.79\% \\
& & & Avg. softmax 
& 58.29\% & 6.20 & 48.68\% & 68.42\% \\ 
\hline

\multirow{6}{*}{Mixed-Language} & \multirow{2}{*}{HE+EN} & \multirow{2}{*}{HE} & Majority vote 
& 74.61\% & 2.06 & 72.37\% & 77.63\% \\
& & & Avg. softmax 
& 73.82\% & 2.36 & 68.42\% & 76.32\% \\
\cline{2-8}
& \multirow{2}{*}{HE+EN} & \multirow{2}{*}{EN} & Majority vote 
& 79.34\% & 3.29 & 73.68\% & 82.89\% \\
& & & Avg. softmax 
& 79.21\% & 3.15 & 73.68\% & 81.58\% \\
\cline{2-8}
& \multirow{2}{*}{HE+EN} & \multirow{2}{*}{HE+EN} & Majority vote 
& 75.13\% & 2.52 & 71.05\% & 78.95\% \\
& & & Avg. softmax
& 75.13\% & 2.10 & 71.05\% & 77.63\% \\
\hline

\end{tabular}
}
\end{center}
\end{table}
%------------------------------

\subsection{Human Test Results}
In order to compare our results with those of human examiners, we developed a mobile application that tests the accuracy of humans on the same task. The application was distributed among 153 females and 147 males; each of the 300 participants received 15 Hebrew handwritings and 15 English handwritings chosen at random (from our HEBIU dataset), and was asked to predict the writer's gender of each examined text. The average classification accuracy for English and Hebrew handwritings were 63.6\% and 66.2\%, respectively (both with a standard deviation of 0.13). Females achieved slightly better results than males in both cases. Specifically, they obtained an accuracy of 64.8\% (vs. 62.2\%) for English, and an accuracy of 67.4\% (vs. 65\%) for Hebrew. No correlations between the accuracy and either age group or education level were observed.

\section{Concluding Remarks} \label{furtherWork}
In this paper, we proposed an automatic deep learning scheme for binary gender classification from handwriting images. Specifically, we presented a CNN variant for this task without ``manual'' feature selection/extraction. Our module is relatively simple, yet efficient, in terms of training speed and running time. We considered seven cross-language cases, including training on a Semitic language (Hebrew) and validation on a non-Semitic one (English), and vice versa. Our classification results are comparable to those of previous methods, and are significantly better than those obtained by human examiners on the same dataset.

In addition, we presented a new offline handwriting dataset (the HEBIU dataset), which contains hundreds of labeled handwriting samples in both Hebrew and English, including diverse demographic information.

Our future work will focus on predicting additional attributes of a given writer, e.g., handedness, age group, whether the text is written in the subject's mother tongue, etc. In addition, we plan to apply our approach to other existing handwriting datasets and aim to enlarge our dataset by collecting more handwriting samples, possibly in additional languages.

\bibliographystyle{splncs03}
\bibliography{handwriting-gender}

\begin{thebibliography}{10}
\providecommand{\url}[1]{\texttt{#1}}
\providecommand{\urlprefix}{URL }

\bibitem{ahmed2017improving}
Ahmed, M., Rasool, A.G., Afzal, H., Siddiqi, I.: {I}mproving {H}andwriting
  {B}ased {G}ender {C}lassification {U}sing {E}nsemble {C}lassifiers. {E}xpert
  {S}ystems with {A}pplications  85,  158--168 (2017)

\bibitem{akbari2017wavelet}
Akbari, Y., Nouri, K., Sadri, J., Djeddi, C., Siddiqi, I.: {W}avelet-{B}ased
  {G}ender {D}etection on {O}ff-{L}ine {H}andwritten {D}ocuments {U}sing
  {P}robabilistic {F}inite {S}tate {A}utomata. {I}mage and {V}ision {C}omputing
   59,  17--30 (2017)

\bibitem{al2012quwi}
Al~Maadeed, S., Ayouby, W., Hassa{\"\i}ne, A., Aljaam, J.M.: {QUWI}: {A}n
  {A}rabic and {E}nglish {H}andwriting {D}ataset for {O}ffline {W}riter
  {I}dentification. In: {F}rontiers in {H}andwriting {R}ecognition
  {I}nternational {C}onference on. pp. 746--751. {IEEE} (2012)

\bibitem{al2014automatic}
Al~Maadeed, S., Hassaine, A.: {A}utomatic {P}rediction of {A}ge, {G}ender, and
  {N}ationality in {O}ffline {H}andwriting. {EURASIP} {J}ournal on {I}mage and
  {V}ideo {P}rocessing (1), ~10 (2014)

\bibitem{bandi2005writer}
Bandi, K.R., Srihari, S.N.: {W}riter {D}emographic {C}lassification {U}sing
  {B}agging and {B}oosting. In: {P}roc. 12th Int. {G}raphonomics {S}ociety
  {C}onference. pp. 133--137 (2005)

\bibitem{bouadjenek2014local}
Bouadjenek, N., Nemmour, H., Chibani, Y.: {L}ocal {D}escriptors to {I}mprove
  {O}ff-{L}ine {H}andwriting-{B}ased {G}ender {P}rediction. In: {S}oft
  {C}omputing and {P}attern {R}ecognition, 6th {I}nternational {C}onference of.
  pp. 43--47. {IEEE} (2014)

\bibitem{bouadjenek2015age}
Bouadjenek, N., Nemmour, H., Chibani, Y.: {A}ge, {G}ender and {H}andedness
  {P}rediction from {H}andwriting {U}sing {G}radient {F}eatures. In: {D}ocument
  {A}nalysis and {R}ecognition, 13th {I}nternational {C}onference on. pp.
  1116--1120. {IEEE} (2015)

\bibitem{cha2001priori}
Cha, S.H., Srihari, S.N.: {A} {P}riori {A}lgorithm for {S}ub-{C}ategory
  {C}lassification {A}nalysis of {H}andwriting. In: {D}ocument {A}nalysis and
  {R}ecognition. {P}roceedings. {S}ixth {I}nternational {C}onference on. pp.
  1022--1025. {IEEE} (2001)

\bibitem{djeddi2014lamis}
Djeddi, C., Gattal, A., Souici-Meslati, L., Siddiqi, I., Chibani, Y., El~Abed,
  H.: {LAMIS}-{MSHD}: {A} {M}ulti-{S}cript {O}ffline {H}andwriting {D}atabase.
  In: {F}rontiers in {H}andwriting {R}ecognition, 14th {I}nternational
  {C}onference on. pp. 93--97. {IEEE} (2014)

\bibitem{gattal2018gender}
Gattal, A., Djeddi, C., Siddiqi, I., Chibani, Y.: {G}ender {C}lassification
  from {O}ffline {M}ulti-{S}cript {H}andwriting {I}mages {U}sing {O}riented
  {B}asic {I}image {F}eatures. {E}xpert {S}ystems with {A}pplications  99,
  155--167 (2018)

\bibitem{hull1994database}
Hull, J.J.: {A} {D}atabase for {H}andwritten {T}ext {R}ecognition {R}esearch.
  {IEEE} {T}ransactions on {P}attern {A}nalysis and {M}achine {I}ntelligence
  16(5),  550--554 (1994)

\bibitem{king2000illusory}
King, R.N., Koehler, D.J.: {I}llusory {C}orrelations in {G}raphological
  {I}nference. {J}ournal of {E}xperimental {P}sychology: {A}pplied  6(4),  336
  (2000)

\bibitem{liwicki2007automatic}
Liwicki, M., Schlapbach, A., Loretan, P., Bunke, H.: {A}utomatic {D}etection of
  {G}ender and {H}andedness from {O}n-{L}ine {H}andwriting. In: {P}roc. 13th
  {C}onf. of the {G}raphonomics {S}ociety. pp. 179--183 (2007)

\bibitem{mahmoud2014khatt}
Mahmoud, S.A., Ahmad, I., Al-Khatib, W.G., Alshayeb, M., Parvez, M.T.,
  M{\"a}rgner, V., Fink, G.A.: {KHATT}: {A}n {O}pen {A}rabic {O}ffline
  {H}andwritten {T}ext {D}atabase. {P}attern {R}ecognition  47(3),  1096--1112
  (2014)

\bibitem{IAMOnDB}
Marti, U., Bunke, H.: {T}he {IAM}-{D}atabase: {A}n {E}nglish {S}entence
  {D}atabase for {O}ff-{L}ine {H}andwriting {R}ecognition. {I}nt. {J}ournal on
  {D}ocument {A}nalysis and {R}ecognition  5,  39--46 (2002)

\bibitem{mirza2016gender}
Mirza, A., Moetesum, M., Siddiqi, I., Djeddi, C.: {G}ender {C}lassification
  from {O}ffline {H}andwriting {I}mages {U}sing {T}extural {F}eatures. In:
  {F}rontiers in {H}andwriting {R}ecognition ({ICFHR}), 15th {I}nternational
  {C}onference on. pp. 395--398. {IEEE} (2016)

\bibitem{morera2018gender}
Morera, {\'A}., S{\'a}nchez, {\'A}., V{\'e}lez, J.F., Moreno, A.B.: {G}ender
  and {H}andedness {P}rediction from {O}ffline {H}andwriting {U}sing
  {C}onvolutional {N}eural {N}etworks. Complexity  (2018),
  \url{https://www.hindawi.com/journals/complexity/2018/3891624/}

\bibitem{nair2010rectified}
Nair, V., Hinton, G.E.: {R}ectified {L}inear {U}nits {I}mprove {R}estricted
  {B}oltzmann {M}achines. In: {P}roceedings of the 27th {I}nternational
  {C}onference on {M}achine {L}earning. pp. 807--814 (2010)

\bibitem{shackleton1994european}
Shackleton, V., Newell, S.: {E}uropean {M}anagement {S}election {M}ethods: {A}
  {C}omparison of {F}ive {C}ountries. {I}nternational {J}ournal of {S}election
  and {A}ssessment  2(2),  91--102 (1994)

\bibitem{siddiqi2015automatic}
Siddiqi, I., Djeddi, C., Raza, A., Souici-Meslati, L.: {A}utomatic {A}nalysis
  of {H}andwriting for {G}ender {C}lassification. {P}attern {A}nalysis and
  {A}pplications  18(4),  887--899 (2015)

\bibitem{youssef2013automated}
Youssef, A.E., Ibrahim, A.S., Abbott, A.L.: {A}utomated {G}ender
  {I}dentification for {A}rabic and {E}nglish {H}andwriting  (2013)

\end{thebibliography}
\end{document}